\documentclass[letterpaper, 10 pt, conference]{ieeeconf}

\IEEEoverridecommandlockouts 

\makeatletter
\let\NAT@parse\undefined
\makeatother
\usepackage[numbers,sort&compress]{natbib}
\usepackage{placeins}

\usepackage{amsmath} 
\usepackage{amssymb} 
\usepackage{bm}
\usepackage{graphicx}
\usepackage{mathtools}

\usepackage{siunitx}
\usepackage{tikz}
\usetikzlibrary{shapes}
\usepackage{multirow}
\usepackage{makecell}
\usepackage{cuted}
\usepackage{textcomp}
\usepackage{stfloats}
\usepackage{url}
\usepackage{verbatim}
\usepackage{cite}
\usepackage{gensymb}
\usepackage{multirow}
\usepackage{booktabs}

\usepackage{graphicx}
\usepackage{subcaption} 

\usepackage{subcaption}
\usepackage{tablefootnote}

\usepackage{bm}
\usepackage{algorithm}
\usepackage[noend]{algpseudocode}
\usepackage{multirow}
\usepackage{subcaption}
\usepackage{setspace}
\usepackage{fancyhdr}
\usepackage{hyperref} 
\usepackage{cleveref}

\usepackage{xcolor} 

\renewcommand{\baselinestretch}{0.99}
\Crefformat{figure}{#2Fig.~#1#3}
\Crefmultiformat{figure}{Figs.~#2#1#3}{ and~#2#1#3}{, #2#1#3}{ and~#2#1#3}
\Crefformat{equation}{#2Eq.~#1#3}

\usepackage{xspace}
\makeatletter
\DeclareRobustCommand\onedot{\futurelet\@let@token\@onedot}
\def\@onedot{\ifx\@let@token.\else.\null\fi\xspace}
\makeatother

\definecolor{forestgreen}{RGB}{34,139,34}
\definecolor{orange}{RGB}{255, 191, 0}

\definecolor{lightblue}{HTML}{9BB7D4} 
\definecolor{light2blue}{HTML}{92A8D1} 
\definecolor{aqua}{HTML}{7BC4C4} 
\definecolor{turquoise}{HTML}{53B0AE} 
\definecolor{iris}{HTML}{5A5B9F} 

\definecolor{darkyellow}{HTML}{F0C05A} 
\definecolor{yellow}{HTML}{F8D948} 

\definecolor{pink}{HTML}{D94F70} 
\definecolor{lightpink}{HTML}{F7CAC9} 
\definecolor{warmpink}{HTML}{FF6F61} 
\definecolor{darkpink}{HTML}{C74375} 
\definecolor{magenta}{HTML}{BB2649} 

\definecolor{orange}{HTML}{E2583E} 
\definecolor{lightorange}{HTML}{FEBE98} 
\definecolor{orangered}{HTML}{DD4124} 

\definecolor{green}{HTML}{009473} 
\definecolor{grassgreen}{HTML}{88B04B} 

\definecolor{redbrown}{HTML}{955251} 
\definecolor{darkred}{HTML}{9B1B30} 

\definecolor{purple}{HTML}{6968AC} 
\definecolor{darkpurple}{HTML}{5F4B8B} 
\definecolor{lightpurple}{HTML}{B163A3} 

\definecolor{grey}{HTML}{959A9C} 
\definecolor{beige}{HTML}{DECDBE} 

\title{\LARGE \bf Learning Dexterous Grasping from Sparse Taxonomy Guidance}

\author{Juhan Park$^{1}$, Taerim Yoon$^{1}$, Seungmin Kim$^{1}$, Joong-Gil Kim$^{2}$, Wontae Ye$^{2}$, Jeongeun Park$^{3}$, Yoonbyung Chai$^{1}$,\\ Geonwoo Cho$^{2}$, Geunwoo Cho$^{2}$, Dohyeong Kim$^{4}$, Kyungjae Lee$^{5}$, Yong-Jae Kim$^{2,6}$, and *Sungjoon Choi$^{1,7}$ 
\thanks{*: corresponding author.
$^{1}$Juhan Park, Taerim Yoon, Seungmin Kim, Yoonbyung Chai, and Sungjoon Choi are with the Department of Artificial Intelligence, Korea University, Seoul, Korea. $^{2}$Joonggil Kim, Wontae Ye, Geonwoo Cho, Geunwoo Cho, and Yong-Jae Kim are with the Korea University of Technology and Education, Cheonan, Korea. $^{3}$Jeongeun Park is with Naver AI Lab, Korea. $^{4}$Dohyeong Kim is with the Seoul National University, Seoul, Korea. $^{5}$Kyungjae Lee is with the Department of Statistics, Korea University, Seoul, Korea. $^{6}$Yong-Jae Kim is also with WIRobotics, Korea. $^{7}$Sungjoon Choi is also with RLWRLD, Korea.}
\thanks{This work was supported by Institute of Information \& Communications Technology Planning \& Evaluation (IITP) grant funded by the Korea government (MSIT) (No. RS-2019-II190079, Artificial Intelligence Graduate School Program(Korea University), 50\%) and (No. RS-2024-00336738, Development of Complex Task Planning Technologies for Autonomous Agents, 50\%). The project page is \href{https://juhanpark.com/GRIT-project-page/}{\texttt{Grit.github.io}}.}}

\renewcommand{\headrulewidth}{0pt} 

\begin{document} 
    \maketitle 
    \thispagestyle{fancy}
\fancyhf{} 
\renewcommand{\headrulewidth}{0pt} 


\begin{abstract}
Dexterous manipulation requires planning a grasp configuration suited to the object and task, which is then executed through coordinated multi-finger control.
However, specifying grasp plans with dense pose or contact targets for every object and task is impractical.
Meanwhile, end-to-end reinforcement learning from task rewards alone lacks controllability, making it difficult for users to intervene when failures occur.
To this end, we present \textbf{GRIT}, a two-stage framework that learns dexterous control from \emph{sparse taxonomy guidance}. 
GRIT first predicts a taxonomy-based grasp specification from the scene and task context. Learning Dexterous Grasping from Sparse Taxonomy Guidance 
Conditioned on this sparse command, a policy generates continuous finger motions that accomplish the task while preserving the intended grasp structure.
Our result shows that certain grasp taxonomies are more effective for specific object geometries. 
By leveraging this relationship, GRIT improves generalization to novel objects over baselines and achieves an overall success rate of 87.9\%.
Moreover, real-world experiments demonstrate controllability, enabling grasp strategies to be adjusted through high-level taxonomy selection based on object geometry and task intent.
\end{abstract}

\section{Introduction}

Dexterous manipulation requires the ability to select and execute appropriate hand movements in accordance with the task context. 
Humans naturally exhibit this capability by flexibly adapting their grasp strategies to the context, including the object’s geometry and the intended use~\citep{feix2015grasp, cutkosky1989grasp, santello1998postural}.
For a deformable object such as a sponge, humans may initially adopt a tripod grasp, which provides precise control of the thumb, index, and middle fingers for stable handling. 
As the task shifts to squeezing out water, humans transition to a power grasp, enabling stronger force generation and whole-hand engagement. \textbf{}
Moreover, humans can switch to alternative grasp strategies when a chosen grasp proves ineffective, demonstrating \emph{controllability} in manipulation.

Developing such a system poses three key challenges.
First, the system must generate appropriate grasp configurations, comprising a discrete taxonomy and a wrist orientation that are suitable for the object geometry and task context~\citep{feix2015grasp, cutkosky1989grasp, song2025overview}.
Second, the policy must translate this sparse grasp configuration into consistent finger-level control despite the absence of explicit motion trajectories~\citep{christen2022d, huang2025fungrasp}.
Finally, the grasp strategy should be controllable through an external command rather than being fixed~\citep{handa2020dexpilot}.



One possible approach is to use large-scale object-grasp data to learn manipulation policies from demonstrations~\citep{an2025dexterous, zhang2024graspxl, chen2025dexonomy}. However, collecting demonstrations covering the combinatorial space of object geometries and task contexts is costly. Instead, we aim to learn from sparse supervisory signals without requiring explicit motion demonstrations. Specifically, the policy learns to use grasp taxonomies as high-level guidance for generating detailed finger motions.

To this end, we propose \textbf{GRIT} (\textbf{G}rasp \textbf{R}einforcement with \textbf{I}ntended \textbf{T}axonomies), a two-stage framework that infers an appropriate grasp configuration and then executes dexterous grasping under sparse grasp taxonomy guidance.
To achieve this, we construct a structured library of grasp taxonomies encompassing a diverse range of human finger-engagement patterns~\citep{feix2015grasp}. This allows the framework to incorporate representative grasp types, providing a versatile foundation for learning dexterous grasping.
In particular, the key contribution of GRIT is learning a control policy conditioned on these grasp configurations, thereby bridging high-level grasp intent and low-level finger execution.

\begin{figure}[t!]
    \centering
    \includegraphics[width=0.95\linewidth]{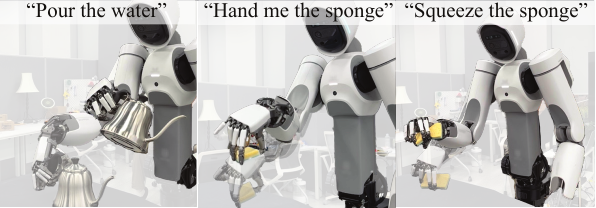}
    \vspace{-5pt} 
    \caption{Our framework selects appropriate grasp taxonomies based on object geometry and task context, and executes them through a taxonomy-conditioned control policy.}    \label{fig:teaser}
    \vspace{-10pt} 
\end{figure}

We conduct extensive experiments to evaluate our framework.
We first evaluate generalization to novel objects in dexterous grasping tasks and show that taxonomy-guided planning and conditioned control outperform baseline methods~\citep{zhang2025robustdexgrasp, zhang2024graspxl}.
Next, we analyze our system’s ability to generate grasp configurations tailored to the object geometry and assess their impact on task performance.
We further examine how faithfully the taxonomy-conditioned policy adheres to the prescribed grasp specification during execution.
Finally, real-world experiments demonstrate controllable dexterous grasping via grasp taxonomies aligned with task intention and object geometry, as illustrated in Fig.~\ref{fig:teaser}.

Our contributions are summarized below.
\begin{itemize}
    \item We propose GRIT, a two-stage framework that generates grasp taxonomies aligned with the context and performs dexterous grasping under sparse taxonomy guidance.
    
    \item We empirically show that grasp taxonomy effectiveness depends on object geometry, and leveraging this improves dexterous grasping on novel objects.

    \item We demonstrate controllability in real-world experiments by adapting grasp taxonomies according to task intention and object geometry.
\end{itemize}





%

\section{Related Work}
Dexterous manipulation is commonly decomposed into \emph{grasp generation and control}~\citep{song2025overview}.
Our approach follows this line of research by structuring the decomposition around grasp taxonomies as an intermediate representation.
From another perspective, our system focuses on generating motor control with sparse taxonomy guidance, addressing the appropriate \emph{level of autonomy} that balances human controllability and autonomous execution.
In the following sections, we review related work from these two perspectives.




\subsection{Grasp Generation and Control}

Humans adapt grasp configurations across diverse object–task combinations, making the joint problem of grasp selection and execution central to dexterous manipulation research~\citep{song2025overview}.

One line of work leverages human demonstrations to bypass explicit grasp generation.
For instance, \citet{herzog2012template} retrieves grasp configurations based on local shape similarity, while \citet{chen2022dextransfer} retargets demonstrations to train multi-finger control policies.
Although these approaches simplify structural modeling, they often struggle to generalize to diverse and novel objects.
Recent studies instead learn grasp strategies from large-scale datasets.
For example, \citet{zhang2024graspxl} learns grasp behaviors from large-scale annotations indicating valid and invalid contact regions, while \citet{chen2025dexonomy} synthesizes grasp configurations using hand templates aligned with object geometry.
However, covering the combinatorial space of object–task variations remains costly.

To address this challenge, several works decompose manipulation into high-level grasp planning and grasp-conditioned control.
For example, \citet{christen2022d} generates grasp taxonomies and interaction locations to condition low-level policies, and \citet{zhang2025dora} uses object affordance maps to produce grasp candidates and constrain policy learning.
Our method follows this decoupled structure but additionally incorporates user intention into grasp planning.

\subsection{Level of Autonomy in Dexterous Manipulation}

Dexterous manipulation systems can also be categorized by their level of autonomy.
Fully autonomous approaches, such as end-to-end reinforcement learning methods~\citep{akkaya2019solving, chen2022towards, chen2022system, zhang2025robustdexgrasp}, learn a direct mapping from observations to actions to optimize task performance.
While effective across many tasks, these systems provide limited controllability after training, making it difficult to adjust behaviors during deployment.
At the other extreme, teleoperation-based systems provide fine-grained control by directly specifying hand motions.
For example, \citet{handa2020dexpilot} tracks human hand poses using multi-view vision and retargets them to a robotic hand, while \citet{zhang2025doglove} uses a wearable glove to map human finger and wrist motions to robot joints.
Although these methods enable precise joint-level control, they require continuous human involvement.

To balance automation and controllability, recent work adopts semi-autonomous frameworks that separate high-level grasp specification from low-level control execution.
For instance, \citet{christen2022d} conditions control policies on externally provided grasp poses, and \citet{huang2025fungrasp} infers grasp poses from RGB observations to serve as high-level guidance.
However, these methods rely on dense grasp representations, such as continuous hand poses or object-specific contact targets.

Our approach adopts a similar semi-autonomous structure consisting of a vision-language model (VLM)~\citep{gemini3pro2025} based grasp planner and a grasp-conditioned controller.
Unlike prior work, we represent grasp intent using a sparse, discrete set of human grasp taxonomies rather than continuous pose specifications.
This allows detailed finger motions to be learned by the low-level policy while enabling flexible high-level guidance across tasks.


\section{Problem Formulation}




Dexterous grasping requires selecting an appropriate grasp configuration according to the scene and task context, and executing it through coordinated finger motions. 
We formulate this process as a policy learning problem where a robot receives high-level grasp intent and generates continuous control actions to accomplish the manipulation.

Let $\mathcal{C}$ denote the global task context, $\mathcal{C} = \{I, T\}$, where $I$ represents an RGB-D observation of the scene and $T$ denotes a task description specifying the user's intent.

The robot interacts with the environment through a control policy $\pi$ that produces actions based on the current state and the task context. 
The objective is to learn a policy that maximizes the expected cumulative reward

\begin{equation}
\pi^* = 
\arg\max_{\pi}
\mathbb{E}_{\tau \sim \pi}
\left[
\sum_{t=0}^{T}
\gamma^t r_t(s_t, a_t, g)
\right],
\end{equation}
where $s_t$ denotes the robot state at time $t$, $a_t$ is the control action, $r_t$ is the reward function encouraging successful task completion and stable grasping, and $\gamma$ is the discount factor.

\begin{figure*}
    \centering
    \vspace{0.3cm}
    \includegraphics[width=0.99\linewidth]{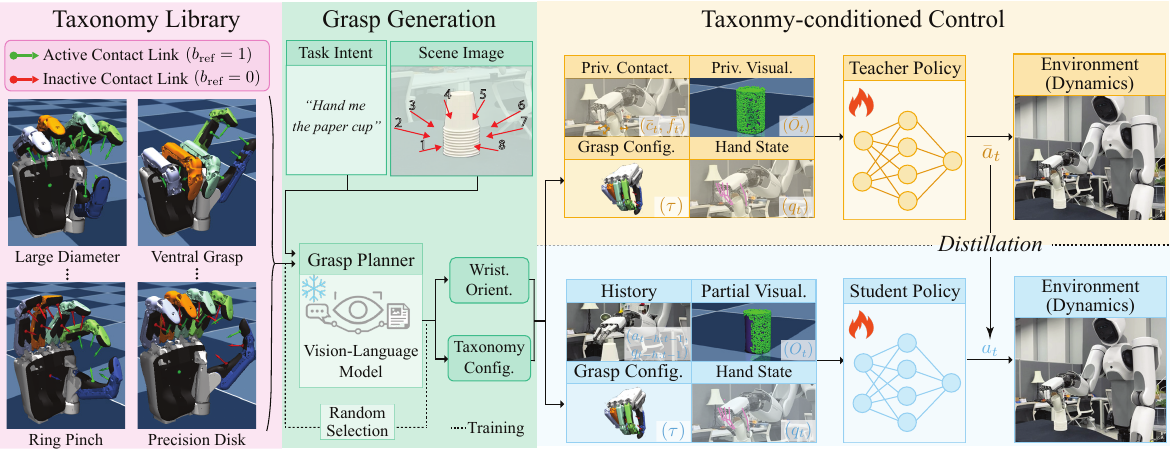}
    \caption{The proposed framework consists of three main components.
    1) A taxonomy library providing canonical grasp templates defined by reference hand configurations and contact structures.
    2) A grasp generation module that randomly samples a taxonomy during training, while at inference, a vision–language model selects one from the scene and task description.
    3) A taxonomy-conditioned control policy learned via teacher–student distillation, where the teacher uses privileged information and the student relies on action–state history and partial visual observations.}
    \label{fig:overview}
    \vspace{-8pt}
\end{figure*}





\subsection{Decoupling Grasp Generation and Control} \label{sec: decouple}
Directly learning a policy that maps the task context $\mathcal{C}$ to dexterous actions is challenging due to the high dimensionality of both perception and control. 
Instead, we decompose the problem into two stages: high-level grasp selection and low-level conditioned control.

First, we define a taxonomy $\tau_i \in \mathcal{T}$ as a grasp template.
Let $\tilde{q} \in \mathbb{R}^D$ denote the reference joint configuration of the hand, and 
$\tilde{b} \in \{0,1\}^L$ denotes a binary mask indicating the engagement of each of the $L$ hand and palm links ($1$: active, $0$: resting).
Let $\tilde{p}, \tilde{n} \in \mathbb{R}^{L \times 3}$ denote the reference contact positions and corresponding surface normals in the local hand frame, providing object-agnostic geometric guidance.
Thus, $\tau_i = \{\tilde{q}, \tilde{b}, \tilde{p}, \tilde{n}\}$.

Using this template, a planner selects a grasp configuration 
$g = \pi_{\text{plan}}(\mathcal{C})$, 
where $g = (\tau, \bar{w}_w)$ consists of a grasp taxonomy $\tau$ selected from the predefined taxonomy set $\mathcal{T}$ and a target wrist orientation $\bar{w}_w$.

Given the selected grasp configuration $g$, the taxonomy-conditioned controller produces continuous control actions conditioned on the current state and the grasp specification,
$a_t = \pi_{\text{control}}(s_t, g)$.
This decomposition enables the use of large perception models for grasp selection while focusing learning on the control policy, formulated as
\begin{equation}\label{eq:controller_objective}
J(\pi) =
\mathbb{E}_{\pi}
\left[
\sum_{t=0}^{T}
\gamma^t r_t(s_t, g)
\right].
\end{equation}

Note that the grasp configuration $g$ provides only sparse guidance, making stable dexterous control challenging to learn.
To address this, we incorporate reward design and feature-extraction strategies to facilitate policy learning under sparse-taxonomy conditioning. 
Further details are provided in~\Cref{sec: controller}.

\subsection{State and Action Representation}
At each timestep $t$, the controller observes a state $s_t = \{s_t^h, s_t^o\}$ and produces an action $a_t$, where $s_t^h$ denotes the hand proprioceptive and contact-aware state and $s_t^o$ represents the object state.

Let $q \in \mathbb{R}^D$ denote the current joint configuration and $\bar{q}^{-}$ the target joint configuration predicted by the policy at the previous timestep. 
The hand state is defined as 
$s^h = [q, \bar{q}^{-} - q, v_w, \omega_w, v_o, \omega_o, \mathbf{c}, \mathbf{f}]$, 
where $(v_w, \omega_w)$ and $(v_o, \omega_o)$ denote the linear and angular velocities of the wrist and object, respectively, and $\mathbf{c}, \mathbf{f}$ represent contact indicators and contact forces. 
The object state $s_t^o$ is represented by a partial point cloud $O \in \mathbb{R}^{n \times 3}$ obtained from a single RGB-D camera and expressed in the wrist frame.

The policy predicts desired joint and wrist configurations $(\bar{q}, \bar{p}_w, \bar{w}_w)$. 
The joint target $\bar{q}$ is tracked using a PD controller, while the wrist pose is enforced by applying corrective forces that reduce the pose error following Baumgarte stabilization~\citep{baumgarte1972stabilization}. 
This formulation ensures that wrist control remains compatible with inverse kinematics approaches~\citep{chiaverini1994review} widely adopted in real-world deployment.

The policy outputs an action 
$a_t = [\Delta w_t, \Delta q_t]$, 
where $\Delta w_t \in \mathbb{R}^7$ denotes the delta pose of the wrist (Cartesian displacement and quaternion rotation) and $\Delta q_t \in \mathbb{R}^{D}$ represents the joint displacement for the $D$ degrees of freedom of the hand. 
The target configurations are obtained by accumulating the predicted deltas with the current state as 
$\bar{w}_t = w_t + \Delta w_t$ and $\bar{q}_t = q_t + \Delta q_t$.


\section{Method}
As illustrated in Fig.~\ref{fig:overview}, our framework uses grasp taxonomies to bridge high-level semantic intent and low-level control. 
We first describe how canonical grasp templates are constructed, and then introduce a planner that selects an appropriate taxonomy from the scene and task context. 
Next, we present the feature extraction process used to construct policy observations and the reward design for training the controller. 
Finally, we describe a distillation procedure that enables deployment in the real-world.




\subsection{Taxonomy-Based Grasp Templates}
We construct grasp templates based on the human grasp taxonomy of Feix \textit{et al.}~\citep{feix2015grasp}. 
We select a subset of 30 types, excluding three highly object-specific patterns, and construct grasp templates for robotic manipulation using the representation in Section~\ref{sec: decouple}.
Each template specifies a reference joint configuration, the engaged fingers, and the intended contact structure. 
Contact locations are defined at the center of each fingerpad, while contact directions follow the fingertip surface normals, providing object-agnostic geometric guidance for control.

\subsection{Grasp Selection}
The grasp selection module serves as a high-level planner that predicts the grasp configuration $g=(\tau, \bar{w}_w)$ tailored to the given task $T$ and current scene image $I$. During policy learning, we uniformly sample the grasp taxonomy and target wrist orientation.
In inference, we utilize a Vision-Language Model (VLM) to perform zero-shot grasp selection. To overcome the difficulty the VLM faces in reasoning about 3D spatial relationships from 2D images~\citep{liu20253daxisprompt}{, we sample a candidate wrist pose} and overlay a 3D coordinate axis directly onto the scene image to represent the potential target direction toward the object, as illustrated in the \textit{Grasp Generation} module in Fig.~\ref{fig:overview}. 
By processing the augmented image and the task description (e.g., "Pick up the sponge to squeeze it") simultaneously, the VLM selects the appropriate taxonomy $\tau$. Detailed evaluation is provided in \Cref{sec: planner evaluation}.

\subsection{Feature Extraction}
To condition the policy on the grasp configuration and object geometry, we construct features from the grasp command $g$ and the partial object point cloud $O$.

First, we define the grasp command feature
$o^g = [\tilde{q} - q, \tilde{b}, \bar{w}_w]$,
where $\tilde{q}$ and $\tilde{b}$ denote the reference joint configuration and link engagement mask of the selected taxonomy, and $\bar{w}_w$ represents the target wrist orientation.

From the point cloud $O$, we extract geometric features describing the spatial relationship between the hand and the environment.
We compute the wrist–object displacement $x_{\text{rel}} \in \mathbb{R}^3$, where the object position is defined as the mean of the partial point cloud.
Following \citep{zhang2025robustdexgrasp}, we further compute distance-based features, including the displacement from each hand link to the nearest object surface point and the distance to the table, denoted by $d_{\text{hand}}$ and $d_{\text{table}}$.
Finally, we encode local object geometry using a wrist-centered Basis Point Set (BPS)~\citep{prokudin2019efficient}.
The effectiveness of this representation is evaluated in \Cref{sec: ablation}.

\subsection{Taxonomy-condition Control Policy} \label{sec: controller}
To learn dexterous grasping from sparse taxonomy guidance, the controller must simultaneously satisfy several objectives: (1) approaching the object reliably, (2) interacting with it according to the selected grasp taxonomy, and (3) avoiding unstable or physically invalid behaviors such as unintended collisions or excessive motion. Balancing these objectives within a single reward formulation is challenging.

Inspired by recent findings \citep{jaeger2025carl} showing that simplified reward structures scale effectively in large-scale reinforcement learning, we adopt a \textbf{multiplicative composite reward}:
\begin{equation}
    r = r_{h}\cdot \alpha_h + r_{o} \cdot \alpha_o - r_\text{pen}
\end{equation}
where $r_h$ and $r_o$ denote the hand-centric and object-centric progressive rewards, $\alpha_h$ and $\alpha_o \in [0,1]$ are multiplicative constraint coefficients that attenuate rewards under undesirable behaviors, and $r_\text{pen}$ represents explicit penalty terms. 
We evaluate this reward formulation in \Cref{sec: ablation} and \Cref{sec: taxonomy SR}.

\subsubsection{Approach Phase and Constraints}
The hand-centric reward $r_h$ guides the approach phase by encouraging the hand to move toward the object surface:
$$
r_h = \exp(-\gamma_r \|x_\text{rel}\|) + \frac{1}{L^{+}} \sum_{l=1}^{L} b_{\text{ref},l} \exp(-\gamma_l \|d_{\text{link},l}\|),
$$
where $\gamma_r$ and $\gamma_l$ control hand-center proximity and link-wise object distance, respectively, and $L^+ = \Sigma b_\text{ref}$ denotes the number of active links.

The multiplicative constraint coefficient is defined as $\alpha_h = \alpha_\text{dir}\alpha_\text{vel}\alpha_\text{act}$, which suppresses unstable behaviors during the approach phase. To elaborate each term, $\alpha_\text{dir} = \exp(-\gamma_{dir} \| d_\text{w} - d_\text{trgt} \|)$ encourages wrist alignment with the target direction $d_\text{trgt}$, $\alpha_\text{vel} = \exp(-\gamma_v (\| v_w \|^2 + \eta \| \omega_w \|^2))$ penalizes excessive linear and angular wrist motion, and $\alpha_\text{act} = \exp(-\gamma_a \| \Delta a_t - \Delta a_{t-1} \|^2)$ regularizes abrupt action changes to encourage smooth control updates.

\begin{table}[t!]
\centering
\vspace{0.2cm}
\footnotesize
\setlength{\tabcolsep}{2pt}
\caption{Summary of Multiplicative Composite Reward Terms in GRIT}
\label{tab:reward_terms}
\begin{tabular}{lll}
\toprule
\textbf{Category} & \textbf{Term} & \textbf{Symbol} \\
\midrule

\multirow{4}{*}{Overall}
& Hand Reward & $r_h$ \\
& Object Reward & $r_o$ \\
& Coefficients & $\alpha_h, \alpha_o \in [0,1]$ \\
& Penalty & $r_{\mathrm{pen}}$ \\

\midrule

\multirow{3}{*}{Approach ($\alpha_h$)}
& Proximity & $x_{\mathrm{rel}}, d_{\mathrm{link},l}$ \\
& Alignment & $\alpha_{\mathrm{dir}}$ \\
& Stability & $\alpha_{\mathrm{vel}}, \alpha_{\mathrm{act}}$ \\

\midrule

\multirow{4}{*}{Grasp ($\alpha_o$)}
& Contact & $n_{t,l}, c_{t,l}, f_{t,l}$ \\
& Obj. Stability & $\alpha_{\mathrm{obj}}=\alpha_{\mathrm{vel}}\alpha_{xy}$ \\
& Adherence & $\alpha_{\mathrm{mimic}}$ \\
& Margins & $\tau_{\mathrm{act}}, \tau_{\mathrm{rest}}$ \\

\bottomrule
\end{tabular}
\vspace{-8pt}
\end{table}



\subsubsection{Grasp Execution and Taxonomy Adherence}
Our framework is designed to encourage both stable object interaction and adherence to the selected taxonomy. 
We achieve this through a multiplicative reward structure with tolerance margins for the selected taxonomy. 
In more detail, the object-interaction reward $r_o$ is defined as
\begin{equation}
r_o = \sum_{l\in L} b_{\text{ref},l}
\left[
\lambda_d (n_{t,l}^{\top} n_{\text{ref},l})
+ \lambda_c c_{t,l}
+ \lambda_i f_{t,l}
\right],
\end{equation}
where $n_{t,l}$ denotes the contact normal at link $l$, and $\lambda_d$, $\lambda_c$, and $\lambda_i$ weight normal alignment, contact status, and impulse magnitude.

To jointly enforce stable object interaction and taxonomy-consistent grasping, we introduce the multiplicative constraint
$
\alpha_o = \alpha_{\text{obj}}\alpha_{\text{mimic}},
$
where $\alpha_{\text{obj}}$ stabilizes object motion and $\alpha_{\text{mimic}}$ enforces taxonomy-consistent hand synergies.

The object stability term is defined as $\alpha_{\text{obj}} = \alpha_{\text{vel}}\alpha_{\text{xy}}$, where
$\alpha_{\text{vel}} = \exp(-\gamma_{ov} (\|v_o\|^2 + \eta_o \|\omega_o\|^2))$ penalizes excessive object motion and
$\alpha_{\text{xy}} = \exp(-\gamma_{xy} \|x_{\text{obj}}^{xy}\|)$ discourages lateral displacement in the grasp frame.

We further define
$
\alpha_{\text{mimic}} = \exp(-\gamma_m \mathcal{L}_{\text{mimic}}),
$
where $\mathcal{L}_{\text{mimic}}$ measures the deviation from the reference taxonomy configuration:
\begin{equation}
\begin{aligned}
\mathcal{L}_{\text{mimic}} =
&\frac{1}{N_{\text{act}}} \sum_{i=1}^{L}
\left(\max(|q_i-q_{\text{ref},i}|-\tau_{\text{act}},0)\right)^2 \\
&+
\frac{1}{N_{\text{rest}}} \sum_{i=1}^{L}
\left(\max(|q_i-q_{\text{ref},i}|-\tau_{\text{rest}},0)\right)^2 .
\end{aligned}
\end{equation}

Importantly, the margins $\tau_{\text{act}}$ and $\tau_{\text{rest}}$ allow controlled deviations from the taxonomy template, enabling flexibility while preserving the intended grasp structure.

\subsubsection{Unintended Interaction Penalties}

We introduce explicit penalties to suppress physically invalid or taxonomy-inconsistent contacts:
\begin{equation}
\begin{aligned}
r_\text{pen} = 
\sum_{l\in L} \Big[
\lambda_{\text{un}}^{c} c^{\text{un}}_{t,l}
+ \lambda_{\text{un}}^{i} f^{\text{un}}_{t,l}
+ (1 - b_{\text{ref},l})
\big(
\lambda_c c_{t,l}
+ \lambda_i f_{t,l}
\big)
\Big].
\end{aligned}
\end{equation}

In detail, $\lambda_{\text{un}}^{c}$ and $\lambda_{\text{un}}^{i}$ weight penalties for unintended contacts and impulse magnitudes. The taxonomy boolean $b_{\text{ref},l}$ indicates an interaction between the object and the $l$th link; contacts on links with $b_{\text{ref},l}=0$ are therefore penalized.

\subsection{Policy Distillation}
To enable real-world deployment, we distill a teacher policy trained with privileged simulation signals into a student policy that uses only accessible observations.
The teacher observes the full object point cloud $\bar{O}_t$ and ground-truth contact states $\bar{c}_t$ and $\bar{f}_t$, whereas the student uses a single-view partial point cloud $O_t$.
We reconstruct the contact signals from the student observations using an LSTM encoder over the past $k=10$ steps of joint states and actions, similar to \citep{zhang2025robustdexgrasp}.
Training follows a curriculum: we first perform DAgger-style behavior cloning~\citep{ross2011reduction} with contact reconstruction and action imitation losses, and then transition to PPO with an asymmetric critic that retains privileged inputs for the critic while the actor relies only on student observations.

\section{Experiments}

In this section, we evaluate the proposed two-stage framework from four complementary perspectives. 
First, we examine its ability to generalize dexterous manipulation to novel objects compared to baseline methods. 
Second, we analyze whether the taxonomy planner successfully captures meaningful object–taxonomy relationships. 
Third, we evaluate how faithfully the taxonomy-conditioned policy adheres to the prescribed grasp specification during execution. 
Finally, we demonstrate the effectiveness of our taxonomy-conditioned framework in real-world manipulation scenarios.

\subsection{Baselines}

\subsubsection{RDG \citep{zhang2025robustdexgrasp}}
To evaluate the effect of taxonomy conditioning, we compare our framework against RDG , which learns grasping behaviors without any explicit grasp-level conditioning. 
RDG does not incorporate taxonomy or high-level grasp specifications, and instead learns grasp motions directly from geometric and contact-based signals. 
The teacher policy takes the previous action $a_{t-1}$, joint angles $q$, contact states $c$, and the hand-centric distance vector $d_{\text{hand}}, d_\text{table}$ as inputs.

The reward is defined as $r = r_{\text{dis}} + r_{\text{contact}} + r_{\text{height}} + r_{\text{reg}}$, where $r_{\text{dis}}$ encourages hand-object proximity, $r_{\text{contact}}$ enforces desired contacts while penalizing undesired collisions, $r_{\text{height}}$ prevents table collision, and $r_{\text{reg}}$ regularizes motion.

\subsubsection{GraspXL \citep{zhang2024graspxl}}
To evaluate the benefit of explicit grasp specification compared to geometric reward shaping, we compare our framework against GraspXL. 
GraspXL does not condition the policy on predefined grasp types, but instead provides indirect guidance about desirable grasp regions via graspable and non-graspable surface annotations. 
The state features include joint angles $q$, PD tracking error $d$, hand and object velocities $(u_h, u_o)$, contact information $(c, f)$, objective errors $(\tilde v, \tilde \omega, \tilde m)$, and distance vectors to graspable and non-graspable surfaces $(l^{+}, l^{-})$. 
These non-graspable surface labels are provided in their official repository and are used during training.

The reward is defined as $r = r_{\text{goal}} + r_{\text{grasp}}$, where $r_{\text{goal}}$ enforces positional and rotational alignment, and $r_{\text{grasp}}$ promotes stable contact, force distribution, anatomical plausibility, and motion regularization.

\subsubsection{GRIT (Ours)}
GRIT differentiates itself through Basis Point Set (BPS)-based geometric features and a multiplicative reward structure that ensures strict taxonomy adherence. We train a single universal policy by randomly sampling 30 taxonomies at each reset, enabling diverse dexterous behaviors within a single model.
Performance is reported via two protocols: Ours (Oracle), which establishes an upper bound by averaging the top-3 taxonomies for each category, and Ours, which utilizes the specific grasp configuration selected by Gemini 3~\citep{gemini3pro2025} during inference.

\subsection{Training Details.}
We train all methods using 30 YCB objects~\citep{calli2015ycb} using PPO~\citep{schulman2017proximal} within the MuJoCo-Warp~\citep{zakka2025mujoco} simulation environment. To maximize data diversity despite the limited object set, we randomly sample the initial wrist rotation relative to the object's canonical frame. Hyperparameters are kept consistent across all methods to ensure a fair comparison. Training is conducted on a workstation equipped with an NVIDIA RTX 5090 GPU.


\subsection{Performance Metric}
\textbf{Success Rate} (unit: $\%$): A trial is considered successful if the robot achieves stable object acquisition by lifting the object at least 5 cm above its initial resting height and maintaining the lifted state for 3 seconds. 

\textbf{Contact Precision} (Prec.): This metric measures the ratio of intended contacts to total object contacts. A high-precision score indicates that the robot manipulates the object using only the designated links, distinguishing between fingertips and other finger segments, without exhibiting semantically inconsistent or arbitrary contacts.

\textbf{Mean Joint Error} (MJE) (unit: rad): Measures the deviation from the reference posture $\mathbf{q}_{\text{ref}}$ with tolerance $\tau_{\text{eval}}=0.3$ during object holding: 
\begin{equation}
\text{MJE} = \frac{1}{D} \sum_{i=1}^{D} (\max(|\text{q}_i - \text{q}_{{\text{ref},i}}| - \tau_{\text{eval}}, 0))^2
\end{equation}
where $\text{q}$ and $\text{q}_{\text{ref}}$ denote the current and reference joint configurations, respectively, and $\tau_{\text{eval}}=0.3$.

\begin{figure}
	\centering
    \vspace{0.3cm}
	\includegraphics[width=.45\textwidth]{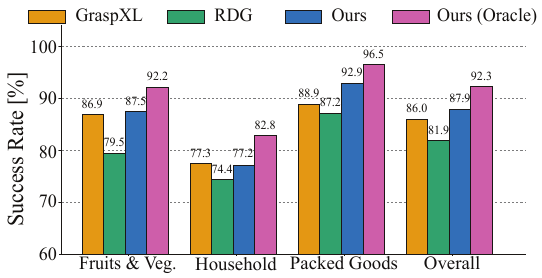}
	\caption{{The object pose is fixed while the wrist orientation is sampled from eight directions around the object. For each direction, 30 trials are performed with randomized initial wrist positions.}}
	\label{figs/success_rate}
    \vspace{-8pt}
\end{figure}

\subsection{Evaluating Generalization to Novel Objects} \label{sec: SR}
We evaluate trained models on the Objaverse dataset \citep{deitke2023objaverse}.
Since the full Objaverse collection is extremely large, we restrict our evaluation 
to the RoboCasa \citep{robocasa2024} subset and reorganize the object labels into three 
categories characterized by their distinct geometric shapes and affordances: \textit{fruit\_vegetable}, \textit{household\_utensil\_tool}, and 
\textit{packed\_goods \& drink \& bread\_food}. 
After filtering, the evaluation set contains 373 objects in total,  
including 135 \textit{fruit\_vegetable}, 
82 \textit{household\_utensil\_tool} objects, and 
156 \textit{packed\_goods \& drink \& bread\_food} objects.

Policy robustness is evaluated with 240 trials per object, consisting of 30 randomized trials for each of 8 approach directions ($45^\circ$ intervals). 
We report the mean success rate across all trials.



As shown in \Cref{figs/success_rate}, our method achieves the best performance among learned policies across all categories. 
Compared to RDG, which does not use taxonomy conditioning, our approach improves the overall success rate from 81.9\% to 87.9\% (+6.0\%). 
It also outperforms GraspXL, which relies on geometric reward shaping, by +1.9\%. 
These results indicate that explicit grasp taxonomy conditioning provides stronger structural guidance for grasp generation.

The gap to oracle selection remains small (87.9\% vs. 92.3\%), suggesting that predicted taxonomies are often near-optimal. 
Performance is lower for \textit{household\_utensil\_tool} objects due to thin or flat geometries that make stable contact difficult, while \textit{packed\_goods \& drink \& bread\_food} objects yield higher success rates due to their regular shapes and larger graspable volumes.

\begin{table}[t!]
\vspace{0.5cm}
\centering
\resizebox{\columnwidth}{!}{
\begin{tabular}{lccc}
\toprule
\multirow{2}{*}{\textbf{Grasp Taxonomy}} 
& \multicolumn{3}{c}{\textbf{Object Category}} \\
\cmidrule(lr){2-4}
& \textbf{Fruit \& Vegetable} 
& \textbf{Household} 
& \textbf{Packed Goods} \\
\midrule
Large Diameter     & 88.38 & 76.71 & 94.19 \\
Medium Diameter    & 88.55 & 76.80 & \textbf{94.31} \\
Small Diameter     & 86.67 & \textbf{78.65} & 93.76 \\
Adducted Thumb     & 86.14 & 75.18 & 93.13 \\
Prismatic 4 Finger & 84.50 & 75.68 & 92.06 \\
Prismatic 3 Finger & 82.75 & 72.60 & 90.56 \\
Prismatic 2 Finger & 82.09 & 72.00 & 89.82 \\
Palmar Pinch       & 84.81 & 74.28 & 91.96 \\
Tip Pinch          & \underline{63.71} & \underline{66.53} & \underline{79.46} \\
Power Sphere       & \textbf{88.78} & 76.16 & 93.74 \\
Tripod             & 81.53 & 71.07 & 89.01 \\
Lateral            & 70.40 & 67.78 & 83.54 \\
Extension Type     & 83.97 & 73.22 & 90.71 \\
Sphere 4 Finger    & 85.97 & 73.38 & 92.74 \\
Sphere 3 Finger    & 77.55 & 70.66 & 87.33 \\
Ventral            & 87.22 & 75.25 & 93.42 \\
\midrule
\makecell{\textbf{Performance Spread} \\ \textbf{(Best--Worst)}} 
& \textbf{25.07} & \textbf{12.12} & \textbf{14.85} \\
\bottomrule
\end{tabular}}
\caption{Success rates (\%) across grasp taxonomies (rows) and object categories (columns). 
\textbf{Bold} indicates the best performance and \underline{underline} indicates the worst.}
\label{tab: obj-SR}
\end{table}

\subsection{Object-Taxonomy Alignment and Sensitivity Analysis} \label{sec: planner evaluation}

To understand the impact of grasp selection, we exhaustively evaluate all 30 taxonomies across three object categories. As shown in \Cref{tab: obj-SR}, success rates vary significantly depending on the taxonomy-object pair, revealing structured preferences.
For instance, \textit{Packed Goods} achieve robust success (94.31\%) with ``medium$/$Large Diameter" grasps. In contrast, \textit{Fruit \& Vegetable} show the highest sensitivity to selection, with a 25.07\% performance gap between the effective ``Power Sphere" (88.78\%) and the struggling ``Tip Pinch" (63.71\%), likely due to the instability of minimal-contact points on highly curved surfaces. These results confirm that object geometry and required contact area necessitate precise taxonomy selection, with power-oriented grasps generally outperforming precision-based ones.

\begin{figure}[t]
\vspace{0.2cm}
\centering
\begin{subfigure}[b]{0.27\linewidth}
    \centering
    \includegraphics[height=3.25cm]{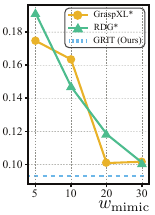}
    \caption{Joint Error ($\scriptstyle\downarrow$)}
    \label{fig:pos_error}
\end{subfigure}
\begin{subfigure}[b]{0.37\linewidth}
    \centering
    \includegraphics[height=3.25cm]{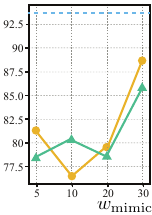}
    \caption{Contact precision ($\scriptstyle\uparrow$)}
    \label{fig:precision}
\end{subfigure}
\begin{subfigure}[b]{0.29\linewidth}
    \centering
    \includegraphics[height=3.25cm]{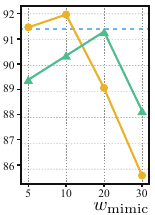}
    \caption{Success Rate ($\scriptstyle\uparrow$)}
    \label{fig:success_rate}
\end{subfigure}

\caption{performance comparison according to the mimic reward weight $w_{\mathrm{mimic}}$. Ours maintains superior performance without hyperparameter tuning. $^*$ Methods retrained with the same taxonomy condition and rewards for a fair comparison.}
\label{fig:mimic_weight}
\vspace{-8pt}
\end{figure}


\subsection{Evaluating Taxonomy Adherence}
\label{sec: taxonomy SR}
We evaluate the policy's adherence to prescribed taxonomies and analyze the trade-off between task success and adherence under different reward formulations.
Baseline methods such as RDG and GraspXL use a standard additive reward formulation. 
To incorporate taxonomy supervision, we follow the same formulation and add the mimic term $\alpha_{\text{mimic}} \in [0,1]$ with a weighting coefficient $w_{\text{mimic}}$, requiring careful tuning to balance taxonomy adherence and task performance.
In contrast, GRIT employs a multiplicative formulation that treats taxonomy adherence as a gating coefficient for the task reward. This design naturally encourages the policy to satisfy the taxonomy constraint without requiring additional hyperparameter tuning.

To evaluate this effect, we retrained baselines with varying $w_\text{mimic}$ and compared them against our fixed formulation.
As shown in Fig.~\ref{fig:mimic_weight}, baseline performance is sensitive to the choice of $w_{\text{mimic}}$, leading to unstable trade-offs between joint error, contact precision, and success rate. 
In contrast, our method maintains consistently strong performance across all metrics without hyperparameter tuning, demonstrating that the multiplicative reward formulation provides a more stable mechanism for enforcing taxonomy adherence while preserving task success.

\begin{figure*}
    \centering
    \vspace{0.3cm}
    \includegraphics[width=1.0\linewidth]{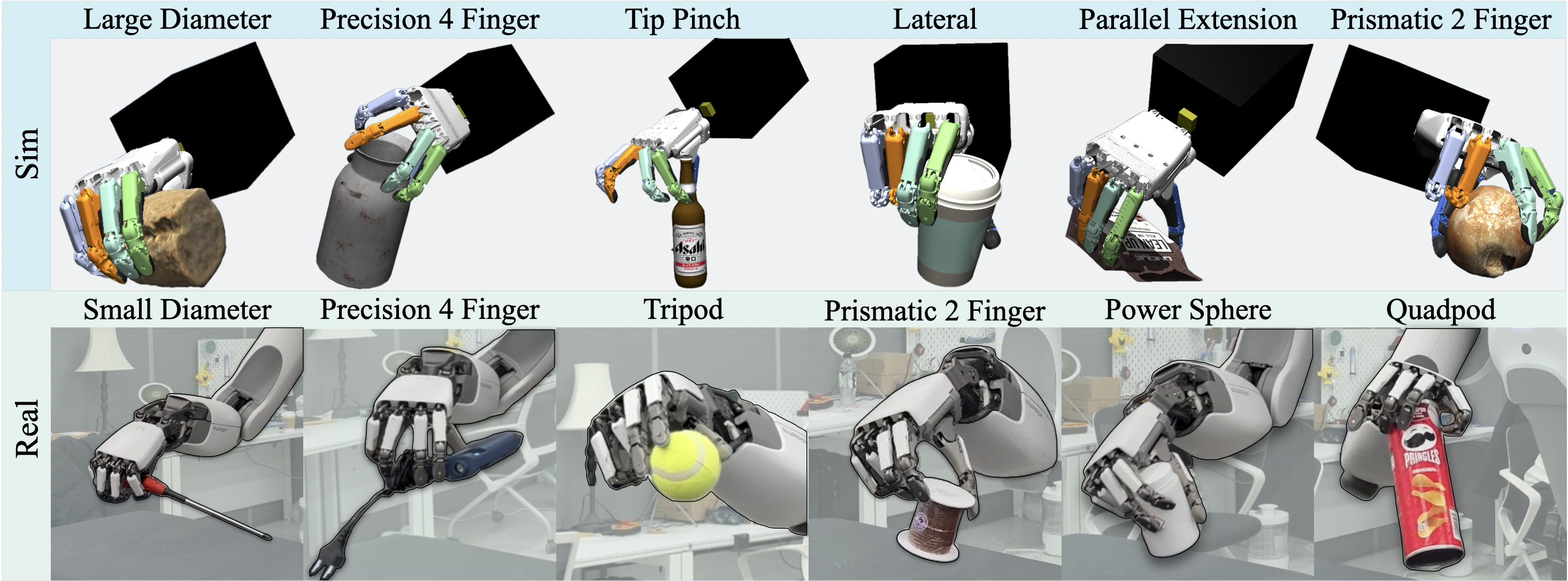}
    \caption{Qualitative examples of generated grasps in simulation and real-world settings.}
    \label{fig:qualitative}
    \vspace{-9pt}
\end{figure*}

\begin{table}[!t]
    \centering
    \caption{Impact of Visual Features and Reward Formulation}
    \label{tab:ablation_study}
    \resizebox{\columnwidth}{!}{
    \begin{tabular}{c|l|cccc|c}
        \noalign{\hrule height 1.5pt}
        \textbf{Finger Usage} & \textbf{Method} & \textbf{SR [\%] $\uparrow$} & \textbf{Prec. $\uparrow$} & \textbf{MJE $\downarrow$} \\
        \hline
        \multirow{3}{*}{2--3 fingers} 
        & W.o BPS        & 76.2 & 0.55  & 0.12 \\
        & Naive sum      & \textbf{80.6} & 0.49  & 0.23 \\
        & \textbf{Ours}  & 79.5 & \textbf{0.63} & \textbf{0.097} \\
        \hline
        \multirow{3}{*}{4--5 fingers} 
        & W.o BPS        & 83.3 & 0.91  & 0.15 \\
        & Naive sum      & 81.1 & 0.90  & 0.39 \\
        & \textbf{Ours}  & \textbf{85.7} & \textbf{0.93} & \textbf{0.19} \\
        \noalign{\hrule height 1.5pt}
    \end{tabular}
    }
    \vspace{-8pt}
\end{table}
\subsection{Ablation Study} \label{sec: ablation}
To further investigate the contribution of each component in our framework—specifically visual features and the multiplicative reward formulation—we conduct an ablation study. We categorize the 30 grasping taxonomies into two groups based on the number of digits involved: 2–3 fingers (e.g., pinch, tripod) and 4–5 fingers (e.g., power sphere, large diameter), as summarized in Table \ref{tab:ablation_study}. 
While a 'naive sum' reward achieves comparable success rates for simple 2–3 finger grasps, it fails to maintain precise finger configurations. In contrast, our multiplicative reward significantly improves adherence, yielding 28.57\% higher Contact Precision and 57.83\% lower joint error, effectively balancing task success with taxonomy adherence. Furthermore, removing BPS results in a consistent performance drop across all metrics, confirming that dense geometric features are essential for the spatial awareness required in high-precision grasping.

\begin{figure}
    \centering
    \vspace{0.3cm}
    \includegraphics[width=0.95\linewidth]{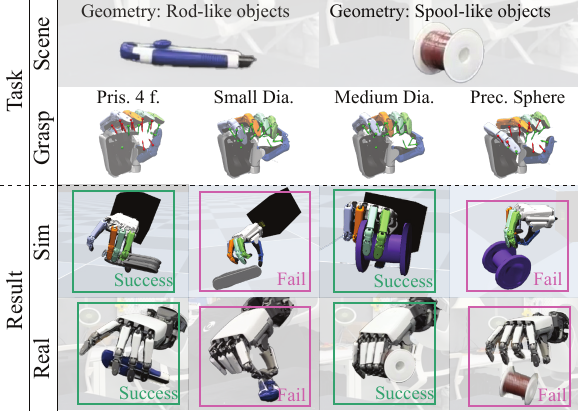}
    \caption{We demonstrate that suitable grasp taxonomies depend on object geometry.
    Across two object geometries, certain taxonomies succeed while others fail in both simulation and real-world experiments.}
    \label{fig:real_experiments}
    \vspace{-8pt}
\end{figure}

\subsection{Real-World Deployment} \label{sec: real-world experiment}
We deploy GRIT on a real-world bimanual humanoid robot using the right-hand and waist joints, where the policy outputs are used as IK targets for the robot wrist. 
As shown in \Cref{fig:qualitative}, GRIT demonstrates successful real-world deployment across a diverse set of grasp taxonomies and objects. Furthermore, we show that the success of each taxonomy depends on both object geometry and task intent, as detailed below.

\textbf{Geometry-Dependent Adaptation}
As shown in \Cref{fig:real_experiments}, the success of each grasp taxonomy depends strongly on object geometry. For rod-like objects such as a knife, the geometry is thin and flat when placed on the table. In this case, diameter-based grasp taxonomies such as \textit{small-diameter} often fail, as the fingers attempt to close around the object but cannot establish stable contact, resulting in empty grasps. In contrast, the \textit{precision 4-finger} taxonomy, which relies primarily on fingertip contacts, successfully grasps the object by pinching the thin geometry. For spool-like objects with a circular cross-section, cylindrical grasp taxonomies such as \textit{medium-diameter} provide stable enclosure and succeed in lifting the object, whereas fingertip-based grasps tend to slip and fail.

\textbf{Task-Specific Selection}
As shown in \Cref{fig:teaser}, GRIT also enables intent-driven control via text prompts beyond geometric fit. When interacting with a sponge, the policy selects a Precision grasp for stable transport with minimal deformation. However, when prompted to ``squeeze,'' it switches to a Power grasp, utilizing the palm and fingers to maximize force generation. This demonstrates that GRIT can adapt grasp strategies not only to object geometry but also to high-level task context.

\section{Conclusions}
We presented \textbf{GRIT}, a two-stage framework for controllable dexterous manipulation that separates grasp taxonomy planning from low-level control. 
A taxonomy planner predicts context-appropriate grasp specifications, and a taxonomy-conditioned policy executes them during manipulation.
Experiments show that taxonomy conditioning improves generalization to novel objects and enables structured object–taxonomy alignment. 
We further demonstrate that the proposed reward design and object representation improve adherence to the specified grasp during execution, and validate the approach in real-world experiments.

Despite these results, several limitations remain. 
The current policy relies on the initial visual observation and lacks real-time feedback to adapt to dynamic scene changes or object displacements. 
In addition, while taxonomy control provides effective high-level guidance, it remains difficult to specify fine-grained targets such as grasping a specific functional part (e.g., a drill handle). 
Future work will focus on integrating reactive visual feedback and more precise control strategies for robust functional manipulation.


\bibliographystyle{IEEEtranN}
\bibliography{root.bib}
\end{document}